\documentclass[letterpaper]{article}
\usepackage{iccc}

\usepackage{times}
\usepackage{helvet}
\usepackage{courier}
\usepackage{multirow}
\usepackage{graphicx}
\usepackage{comment}

\title{MOSAAIC: Managing Optimization towards Shared Autonomy, Authority, and Initiative in Co-creation} 

\author{
  Alayt Issak\thanks{These authors contributed equally to this work, listed in alphabetical order.} \\
  College of Arts, Media and Design\\
  Northeastern University\\
  Boston, MA 02115, USA\\
  issak.a@northeastern.edu
  \And
  Jeba Rezwana\footnotemark[1]\\
  Computer and Information Sciences\\
  Towson University\\
  Towson, MD 21252, USA\\
  jrezwana@towson.edu
  \And
  Casper Harteveld \\
  College of Arts, Media and Design\\
  Northeastern University\\
  Boston, MA 02115, USA\\
  c.harteveld@northeastern.edu
}

\begin{document} 
\maketitle

\begin{abstract}
\begin{quote}

Striking the appropriate balance between humans and co-creative AI is an open research question in computational creativity. Co-creativity, a form of hybrid intelligence where both humans and AI take action proactively, is a process that leads to shared creative artifacts and ideas. Achieving a balanced dynamic in co-creativity requires characterizing control and identifying strategies to distribute control between humans and AI. We define \textit{control} as the power to determine, initiate, and direct the process of co-creation. Informed by a systematic literature review of 172 full-length papers, we introduce MOSAAIC (Managing Optimization towards Shared Autonomy, Authority, and Initiative in Co-creation), a novel framework for characterizing and balancing control in co-creation. MOSAAIC identifies three key dimensions of control: \textit{autonomy}, \textit{initiative}, and \textit{authority}. We supplement our framework with control optimization strategies in co-creation. To demonstrate MOSAAIC's applicability, we analyze the distribution of control in six existing co-creative AI case studies and present the implications of using this framework.

\end{quote}
\end{abstract}

\vspace{-0.5cm}
\section{Introduction}


Human-AI co-creativity refers to the process in which both humans and AI contribute to the creative process to generate creative artifacts or ideas ~\cite{kantosalo_isolation_2014}, producing outcomes that could potentially surpass what either could achieve independently \cite{liapis2014computational}. Unlike traditional creativity support tools designed merely to support human creativity, co-creative AI systems are increasingly being developed as creative partners~\cite{feldman_co-creation_2017,margarido_boosting_2024}. With the rapid rise of Generative AI (GenAI) in co-creative contexts, such as ChatGPT, Runway, and Midjourney, research in co-creativity has gained significant momentum. However, designing effective co-creative AI presents several challenges due to the dynamic nature of human-AI interaction \cite{rezwana2022designing,maher_who_2012}.

There is a growing need to explore the key dimensions of human-AI interaction that shape the distribution of control between users and AI \cite{liu2023use}. Recent advances in co-creative systems, particularly the development of ``Media Foundation Models'' for generating images and videos~\cite{MovieGen}, have further pushed the boundaries of AI's creative capabilities. These advancements along with the complexities of co-creation dynamics have shifted the balance of creative control toward AI \cite{epstein2023art}, raising concerns about maintaining balance in co-creation~\cite{margarido_mi-ccy_2025}.

Traditionally, AI research has focused on building autonomous agents \cite{stuart2003russell}. As a consequence, the role of machines in Human-AI interaction is moving away from the paradigm of ‘AI as a tool’ toward the direction of ‘AI as an agent’ \cite{guzman2020artificial}. In contrast, co-creativity research seeks to design AI systems that function as co-equal creative partners, hence the need for a balanced co-creative system~\cite{kantosalo2020modalities}. Achieving a balanced dynamic in co-creativity requires both humans and AI to maintain creative autonomy while effectively sharing \textit{control}~\cite{muller_interactional_2023}. This balance ensures users retain their sense of control in the process of co-creation~\cite{heer2019agency} and provides harmony to the co-creative process ~\cite{vinchon_artificial_2023}. However, control in co-creativity remains underexplored, as it lacks a clear and unified definition within creativity research. Therefore, it becomes essential to characterize the dimensions of control and develop strategies for control distribution between humans and AI to maintain a ``stable equilibrium'' in co-creativity~\cite{yao_human_2024}. 

In this paper, we aim to address the challenge of characterizing and balancing control in human-AI co-creativity. While much of the existing co-creativity research equates autonomy or agency with control, we adopt a more holistic perspective. We define \textit{\textbf{control}}, as the power to determine, initiate, and direct the process of co-creation. It helps us answer the question of “who is initiating, leading, deciding, navigating... the process of co-creation?” Our research is guided by the following questions, which aim to characterize control and develop strategies for distributing control:

\begin{enumerate}
    \item What are the key dimensions of control in co-creativity? 
    \item How can control be balanced between humans and AI in co-creativity? 
\end{enumerate}

To answer these questions, we conducted a systematic literature review of 172 full-length papers. Informed by the literature review, we propose a framework, Managing Optimization towards Shared Autonomy, Authority, and Initiative in Co-creation (MOSAAIC)\footnote{We use this acronym to allude to a mosaic that is metaphorically used to describe a collection of diverse elements that form a larger whole.}, that characterizes control by identifying the dimensions of control and strategies to balance control in co-creativity. Finally, we use MOSAAIC to analyze the control distribution in six co-creative AI, highlighting current trends and identifying gaps in achieving balanced control in co-creativity. We discuss the implications of the findings along with the value of MOSAAIC.

\section{Background}

Control dynamics between humans and AI is fundamental in co-creation~\cite{margarido_mi-ccy_2025}, but not sufficiently approached as a key concept in the Computational Creativity (CC) literature. Recent work in CC identifies (1) the principle of maximal unobtrusiveness and (2) balance of interjection as driving factors behind achieving balance in co-creation~\cite{lawton_drawing_2023,moruzzi_customizing_2024}. However, these works do not elaborate on control as a mechanism. ~\citeauthor{muller_interactional_2023} (\citeyear{muller_interactional_2023}) define control as the \textit{tactical means of achieving a goal} where it is situated on a hierarchy between initiative (which party is currently acting) and agency (choosing and pursuing a strategic goal); however, this definition lacks a situated understanding of control. We build upon these works and explain how control is encompassed within existing frameworks and roles below. 

\subsection{Frameworks: In Search of Control}

Various works have sought to address control in human-AI co-creation \cite{AI_creativity,rezwana2022designing,kabir_unleash_2024,moruzzi_roles}. These works highlight creativity as the central aim of co-creative systems and wrestle with control dynamics in their plight for balance. They address power imbalances in AI systems, develop mechanisms for adjusting AI contributions, tackle decision-making power over task finalization, and account for societal implications of shifting agency in AI systems. However, while valuable, none of these works foreground control as a key concept. For instance, the Human-AI Co-creation model explains the creative process with AI and the new possibilities brought by AI in each creative phase, but it does not provide guidance on control in the context of co-creation~\cite{AI_creativity}. In our work, we focus on the concept of control and highlight the importance of a balanced control for Human-AI Co-Creativity through a systematic review of the literature. 
We decipher the patterns of control that emerge from works in related disciplines such as Human-Robot Interaction (HRI). 

\subsection{Roles: Defining Control}

Since their advent, human-AI co-creative systems have progressed from tools to partners and agents~\cite[Chapter~8]{manovich_2024}, gradually leading to the design and adoption of AI roles and personas to meet users' needs~\cite{moruzzi_roles}. Different scholars have attempted to identify the roles AI agents can have, however, these works predominantly focus on autonomy~\cite{lubart_how_2005}. \citeauthor{kim2023one} (\citeyear{kim2023one}) categorizes AI into four roles based on autonomy levels: Tools (low in both human and AI autonomy), Servants (high human involvement, low AI autonomy), Assistants (low human involvement, high AI autonomy), and Mediators (high in both dimensions). This has also been formalized in a taxonomy consisting of six levels of autonomy: no AI (level 0), tool (level 1), consultant (level 2), collaborator (level 3), expert (level 4), and agent (level 5)~\cite{morris_levels_2024}. Noting that autonomy is considered to be the key attribute in the formalization of AI roles, in this work, we foreground and define AI roles from the perspective of control 
as it carries a holistic understanding of not only how the process occurs (autonomy), but also how the power to determine, direct, and command are part of the process of co-creation as well. 


\section{Methodology for Systematic Literature Review}

This section outlines the systematic literature review (SLR) procedure used to identify key dimensions of control (RQ1) and strategies to balance control (RQ2). 

\subsection{Identifying Keywords}
We began our SLR by identifying keywords through preliminary research, leveraging the research team's expertise and insights from recent publications in The International Conference on Computational Creativity (ICCC), ACM Conference on Human Factors in Computing Systems (CHI), and ACM Conference on Creativity \& Cognition (C\&C). Through iterative discussions, we selected keywords that capture control dynamics in co-creativity, such as ``control in human-AI co-creation.'' These keywords were iteratively refined by testing them in searching papers and evaluating the relevance of the resulting papers. This iterative approach helped us uncover nuances in keyword selection and optimize our search strategy. While our primary focus was on co-creativity, we discovered valuable tangential work in related fields, such as Human-Robot Interaction (HRI) and the broader AI domain. Consequently, we included papers from both co-creativity and related domains to provide a comprehensive perspective. Finally, collaborative discussions among the authors ensured consensus on the final set of keywords, aligning them with the study's objectives.

\begin{figure}[h]
    \centering
    \includegraphics[width=\columnwidth]{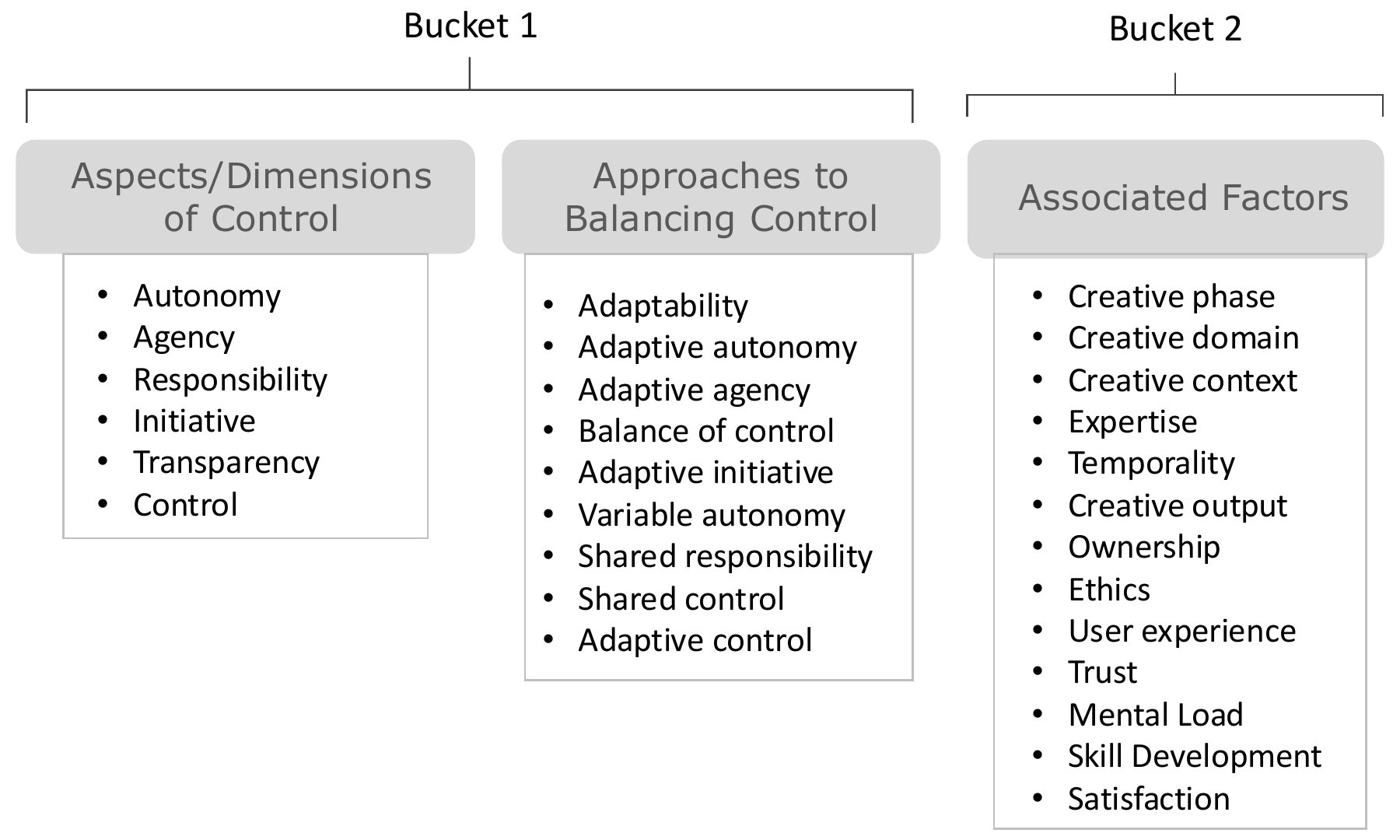}
    \caption{Identified key phrases for Bucket 1 and 2}
    \label{key}
\end{figure}

\begin{figure*}[h]
    \centering
    \includegraphics[width=0.9\textwidth]{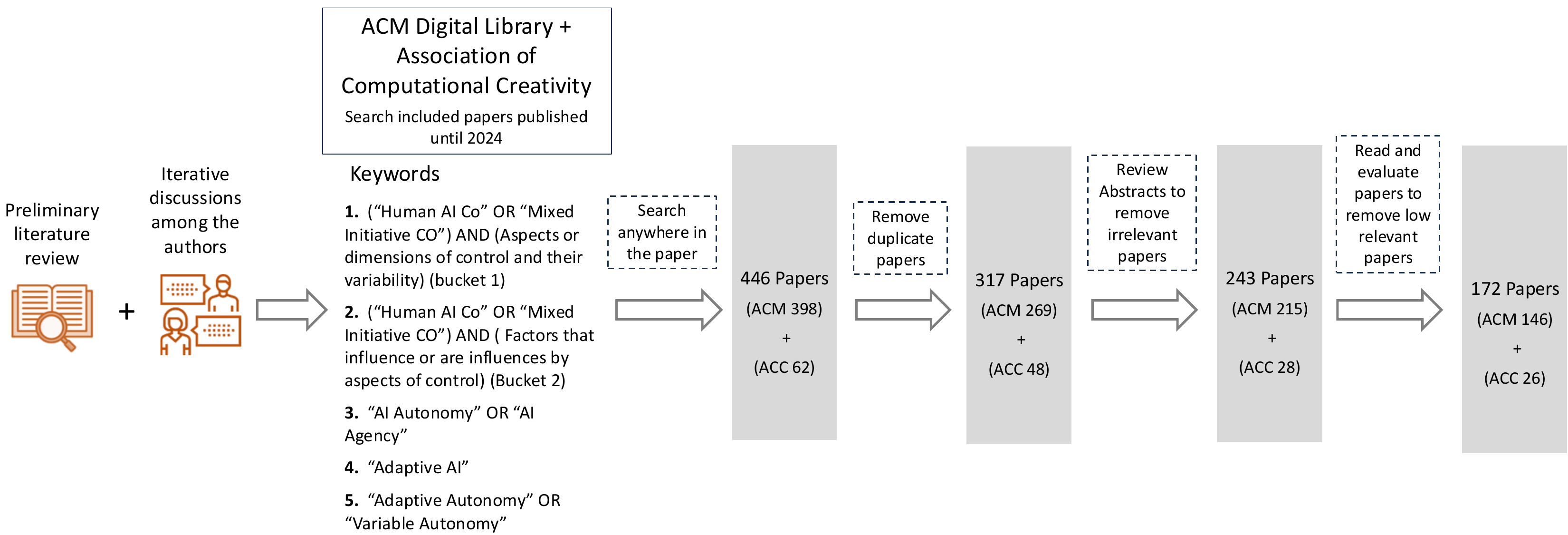}
    \caption{Literature review procedure}
    \label{procedure}
\end{figure*}

We used \textbf{five key phrases} based on our selected keywords to identify relevant papers. To structure our search effectively, we categorized keywords into two primary buckets (Figure~\ref{key}) based on their conceptual role in understanding control dynamics: (Bucket 1) dimensions of control and strategies to balancing control and (Bucket 2) factors that influence or are influenced by control, which explore the broader implications and dependencies of control in co-creative systems. For \textbf{co-creativity research}, we formulated two keyphrases (Figure \ref{procedure}): (1) (Bucket 1 keywords separated by OR) AND (``Human-AI Co'' OR ``Mixed-Initiative Co'') and (2) (Bucket 2 keywords separated by OR) AND (``Human-AI Co'' OR ``Mixed-Initiative Co''). To ensure comprehensive coverage, we used “Human-AI co” as an umbrella term to capture variations such as co-creation, co-creativity, and domain-specific terms like co-dancing. Additionally, since “Mixed-Initiative Co-Creativity” is commonly used in the literature to describe co-creation, we included “Mixed-Initiative Co” to account for variations.

To identify relevant papers in the \textbf{broader AI domain}, we incorporated three additional search key phrases (Figure~\ref{procedure}): (3) ``AI Autonomy'' OR ``AI Agency,'' (4) ``Adaptive AI,'' and (5) ``Adaptive Autonomy'' OR ``Variable Autonomy.'' These terms were selected to encompass diverse perspectives and frameworks related to control and autonomy in AI systems. 


\subsection{Data Collection}

We conducted our literature review (Figure~\ref{procedure}) using two primary databases: the Association for Computing Machinery Digital Library (ACM DL\footnote{https://dl.acm.org}) and the database listed on the Association for Computational Creativity website (ACC\footnote{https://computationalcreativity.net/}). We chose the ACM DL to connect to conferences where HCI, AI, and creativity research are typically published. For searches in the ACM Digital Library, we leveraged its advanced search settings to refine our query. 

We considered the Computational Creativity database (ACC) due to its strong relevance to our research focus. We considered documents published until November 2024. We narrowed our search to full-length papers (e.g., not tutorials or posters) to ensure we considered high-quality peer-reviewed studies. 
Because this database lacks advanced search functionality, we developed a custom Python script to search using our curated keywords.\footnote{https://github.com/jrezwana/MOSAAIC}

After running the keyword-based search, we retrieved 398 papers from the ACM Digital Library (ACM DL) and 62 papers from the Association for Computational Creativity (ACC), totaling 446 papers. Since multiple keyword searches led to duplicate entries, we removed redundant papers, reducing the dataset to 317 unique papers (269 from ACM and 48 from ACC). In the first screening round, the first two authors reviewed the abstracts to assess relevance of the papers to human-AI control dynamics and related aspects. To ensure consistency in the selection process, the authors first reviewed the abstracts (45) from one of the five keyphrase searches together to establish a consensus on inclusion criteria---the paper must directly or tangentially discuss the control dimensions or dynamics or related aspects. They then independently reviewed the abstracts from the remaining four keyword searches, dividing the workload equally. This process resulted in a refined set of 243 papers (215 from ACM and 28 from ACC). In the second round, they conducted a full-text review and further filtered out papers with low relevance, leading to the final corpus of 172 papers, comprising 146 from ACM and 26 from ACC.

\subsection{Data Analysis}
We read the final papers that were highly relevant to our research questions and summarized the literature review on the dimensions of \textit{control} and strategies to balance it in co-creativity using a shared spreadsheet with notes, insights, summaries and scribbles and then organized them into related themes. Our framework was developed in an iterative process of adding, merging, and removing key dimensions of control in human-AI co-creation and strategies to balance control. Through iterative discussion, we developed our framework that presents the dimensions of control and the strategies of balancing control. The framework developed and described below is primarily based on our literature review.

\section{Framework for Managing Optimization towards Shared Autonomy, Authority, and Initiative in Co-creation (MOSAAIC)}

\begin{figure}[h]
    \centering
    \includegraphics[width=\columnwidth]{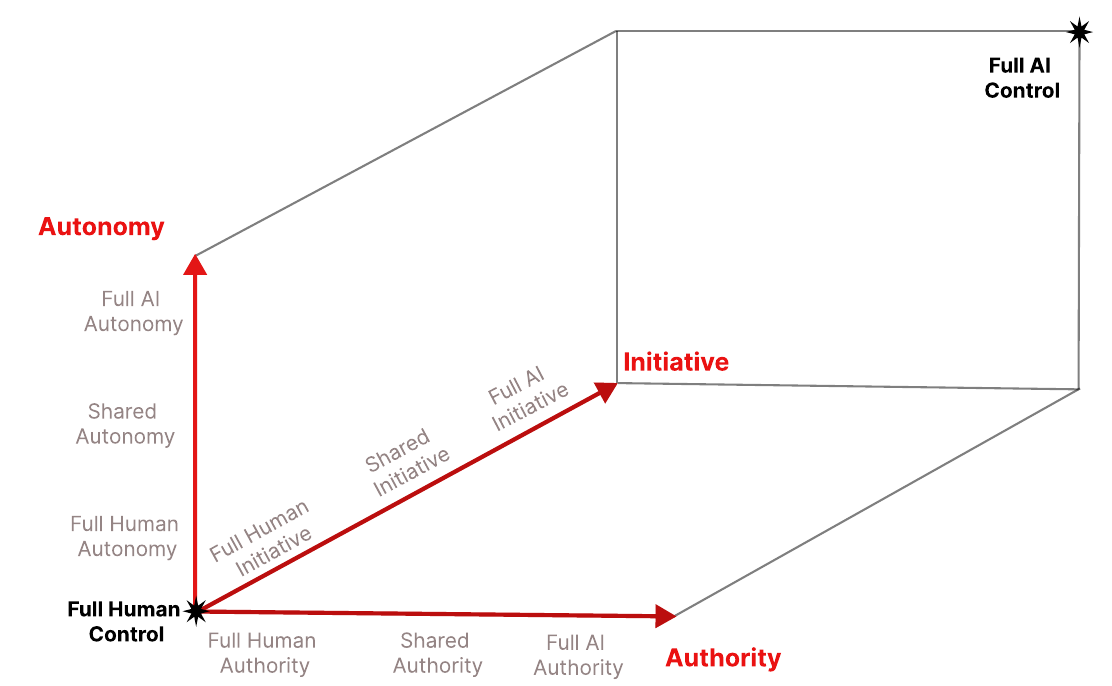}
    \caption{Three dimensions of control in co-creativity}
    \label{framework}
\end{figure}

In this section, we introduce MOSAAIC (Managing Optimization towards Shared Autonomy, Authority, and Initiative in Co-creation), a novel framework that we developed based on our literature review. MOSAAIC characterizes the key dimensions of control in human-AI co-creativity and outlines strategies for achieving a balanced dynamic between humans and AI (Figure~\ref{framework}). The framework defines three key dimensions of control: \textit{Autonomy}, \textit{Initiative}, and \textit{Authority}. \textit{Autonomy} enables an agent to independently choose its creative actions, \textit{initiative} allows it to proactively contribute rather than simply react, and \textit{authority} grants it the power to make decisions and direct the creative process. As illustrated in Figure~\ref{framework}, the dimensions are represented along the axes of a three-dimensional (3D) control space. Each axis reflects a continuum of control, ranging from full human control at one end to full AI control at the other, with shared control in the middle. The corner where all three axes converge at the human-controlled end represents a scenario where humans retain complete control across all dimensions. Conversely, the opposite corner signifies full AI control.

\begin{figure}[h]
    \centering
    \includegraphics[width=0.9\columnwidth]{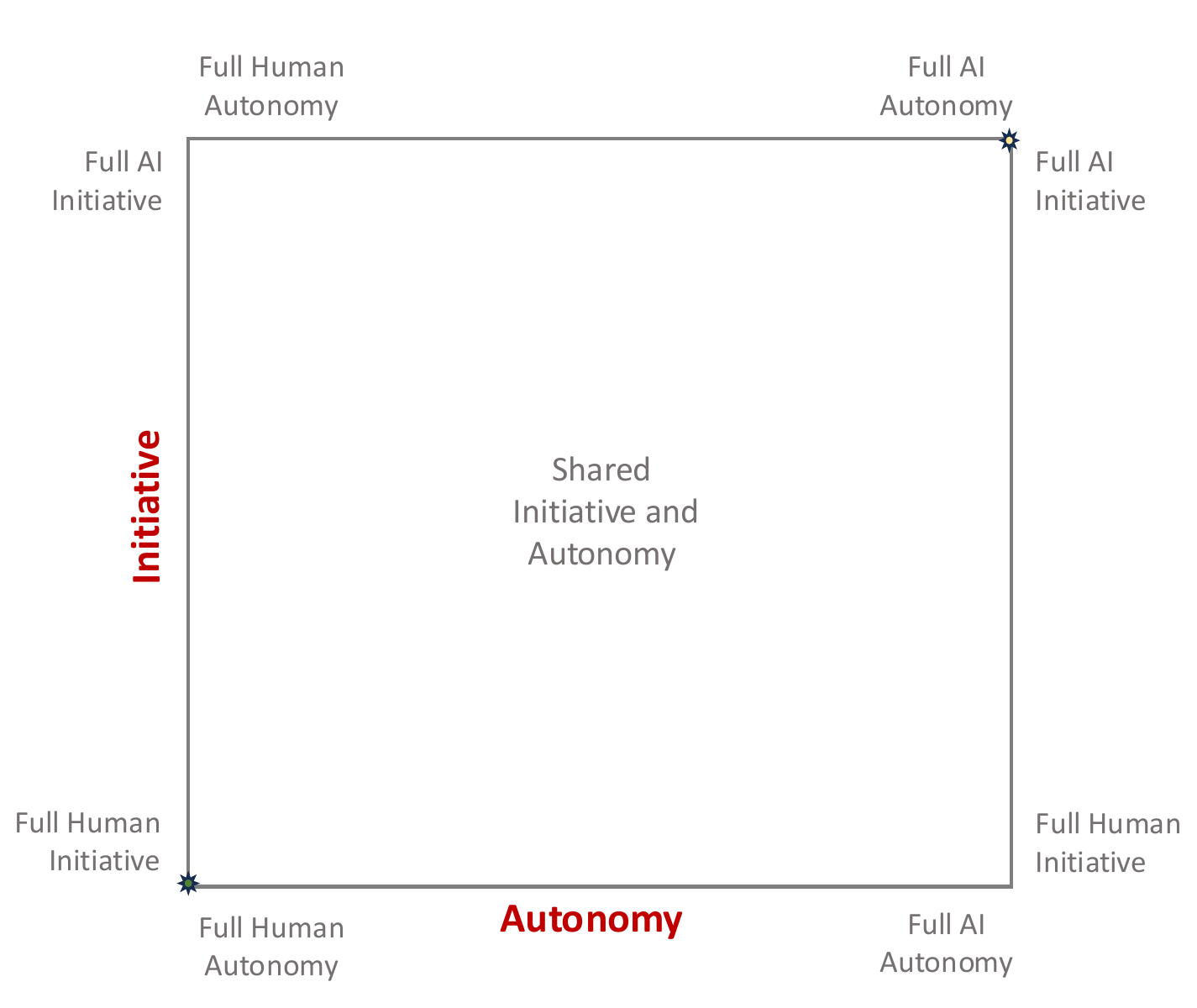}
    \caption{2D plane of autonomy and initiative from the 3D model of MOSAAIC}
    \label{plain1}
\end{figure}

To further break down the 3D model, we present a 2D plane representing a single side of the cube, isolating the relationship between Autonomy and Initiative (Figure \ref{plain1}). In this plane, the horizontal axis represents Autonomy between humans and AI, while the vertical axis represents Initiative. The four corners denote different combinations of high and low levels of human-AI autonomy and initiative. The top-right and bottom-left corners represent scenarios where either humans or AI fully control both autonomy and initiative, while the midpoints along each axis indicate shared autonomy or shared initiative. At the very center of the plane lies balance where both autonomy and initiative are evenly shared between humans and AI---although the right balance would depend on the task to be performed. 

An example of the combination of full-human autonomy and full-AI initiative (top-left corner) could be an improvisational drawing system, where the AI proactively suggests ideas, but the human retains complete control to accept, modify, or ignore them, determining the final creative direction. Conversely, a case of full-human initiative and full-AI autonomy (bottom-right) would be a text-based storytelling co-creative AI, where the human provides a prompt or theme, but the AI fully controls how the story unfolds, with no opportunity for the user to adjust the content once generated. In the following subsections, we discuss the control dimensions and optimization strategies according to MOSAAIC. We begin by defining each dimension and then discussing its role in shaping control dynamics in human-AI co-creativity, drawing insights from the literature. 

\subsection{Three Dimensions of Control}

\subsubsection{Autonomy}
\textit{Autonomy} refers to the ability of the agent to choose its creative action. The MOSAAIC framework identifies three levels of autonomy in co-creativity: \textit{full human autonomy}, where AI acts as a passive assistant responding only to direct instructions; \textit{full AI autonomy}, where the system independently generates creative outputs without human oversight; and \textit{shared autonomy}, where autonomy is distributed between humans and AI, each having a similar degree of freedom to choose their creative actions.

\textbf{Autonomy in Optimizing Control:}
Autonomy is a fundamental aspect of human-AI co-creativity and has been widely discussed in the literature \cite{davis_five_2024}. Research suggests that for co-creativity to be meaningful, AI must possess a certain level of creative autonomy alongside human creators \cite{muller_interactional_2023}. Balancing autonomy between humans and AI is crucial for optimizing control in co-creative systems \cite{moruzzi_customizing_2024}, as these systems exist between two extremes: creativity support tools with low autonomy that primarily assist human creativity and standalone generative AI with full autonomy that does not rely on human input \cite{kantosalo2020modalities}.

Autonomy is closely intertwined with agency, and the two terms are often used interchangeably in HCI research \cite{bennett2023does}. In computational creativity, agency has been conceptualized as the sense of autonomy \cite{pease2023notion}, while \citeauthor{koch_agency_2021} (\citeyear{koch_agency_2021}) defined agency as the ``amount of autonomy” an entity possesses. Given this strong interconnection, MOSAAIC classifies autonomy as one of the key dimensions of control, as agency is a higher level concept that can be defined by autonomy.

The level of autonomy attributed to an AI system is positively correlated with the perceived creativity of its outputs, reinforcing autonomy’s role in human-AI co-creativity \cite{moruzzi2022creative}. Thus, categorizing and applying autonomy effectively in co-creative systems is essential. \citeauthor{davis_five_2024} (\citeyear{davis_five_2024}) propose three levels of AI autonomy—fully autonomous AI, semi-autonomous AI, and user-directed AI—as part of their five-pillar framework for co-creative AI. However, their definition of autonomy combines aspects of both autonomy and initiative, which we distinguish as separate dimensions of control. The classification of autonomy in MOSAAIC, therefore, provides a more nuanced perspective of the degree of autonomy in control distribution.

Co-creativity research emphasizes the importance of shared autonomy, which sits in the middle of the autonomy spectrum in our 3D model, where both humans and AI contribute meaningfully to the creative process \cite{moruzzi_roles}. Shared autonomy has been shown to enhance team performance and user satisfaction \cite{salikutluk2024evaluation}. Effective co-creative partnerships require a balance of AI autonomy with user agency, ensuring that AI enhances—rather than diminishes—human agency and creativity \cite{moruzzi_customizing_2024}. 

\subsubsection{Initiative}
\textit{Initiative} refers to the ability to proactively contribute to the creative process. According to MOSAAIC, \textit{full human initiative} means the AI waits for explicit user input before contributing. \textit{Full AI initiative} allows the system to contribute on its own without human prompting. \textit{Shared initiative} represents an adaptive collaboration where both parties can initiate and respond dynamically, fostering an interactive and evolving creative exchange.

\textbf{Initiative in Optimizing Control:} 
Initiative has been foundational in discerning the human-AI dynamics in co-creativity.
Research on the theory of co-creation highlights that for co-creativity to be mutual, collaborative, and shared, mixed and shifting levels of initiative are essential to the user's sense of control~\cite{wan_it_2024}. Mixed-Initiative Co-Creativity (MICC), where both human and AI systems can initiate contributions and actions, attributes initiative as the fundamental component of achieving co-creativity~\cite{yannakakis_mixed-initiative_2014}. 


In HRI, Mixed-Initiative Control (MIC), where the human and system switch between different levels of initiative by either the human or the system, is a type of shared initiative proposed to optimize control~\cite{mic}. It finds that Human-Initiative (HI) systems outperform systems with single modes of operation (one specific action at a time), whereas Mixed-initiative (MI) systems provide improved performance and improved operator workload~\cite{mic}. MIC also results in better overall performance than giving an agent exclusive control. Specifically, effective MIC arises when agents are capable of making progress toward a goal without having to wait for human input in most circumstances, and when the control interface helps the user manage the progress and intent of several agents~\cite{robot_control}.

Other initiative mechanisms look at turn-taking, where the human and AI take turns at initiating the shared product, as a means of achieving balance~\cite{rezwana2022designing}. They also take the user's initial prompt as a seed and encourage changes, making initiative an iterative process ~\cite{promptcharm}. This entails moving from \textit{full human initiative} to \textit{full AI initiative} in our framework. ~\citeauthor{lobo_when_2024} (\citeyear{lobo_when_2024}) propose AI partners of varying initiative levels to present contextual awareness in the process of co-creation. These levels of initiative are \textit{Leader, Follower,} and \textit{Shifting}. AI partners following human initiative are perceived as warmer and more collaborative, whereas AI leaders are less friendly, although much quicker in execution. However, these outcomes depend on the context (task type) and individual preference of the user, with the potential for personalized MICC systems~\cite{lin_beyond_2023}. Initiative does not exist in a vacuum and operationalizing it to achieve control entails examining other dimensions of control in MOSAAIC. 

\subsubsection{Authority}
\textit{Authority} in human-AI co-creativity refers to the extent of decision-making or directing power that shapes the creative process. It relates to the tactical means of achieving a goal.
According to MOSAAIC, in \textit{full human authority}, AI acts solely as a support tool, while humans retain ultimate authority to direct and decide the tactical means to achieve goals. In contrast, \textit{full AI authority} occurs when AI autonomously determines creative directions without human intervention. \textit{Shared authority} enables negotiation and mutual adaptation, where both human and AI contributions are considered in decision refinement.

\textbf{Authority in Optimizing Control:}
In the literature, the term \textit{control} has often been used to describe what we define as authority within MOSAAIC. In the literature, humans are generally expected to lead co-creative interactions due to their contextual understanding, ethical judgment, and accountability, particularly in complex creative tasks \cite{davis2016co}. In certain contexts, AI authority transfers from humans to AI can optimize efficiency. For example, research on medical imaging suggests that removing humans from the loop can improve accuracy and ethical outcomes \cite{muyskens2024can}. Similarly, some argue that AI should be granted epistemic authority in specific domains where its accuracy surpasses human expertise \cite{ferrario2024experts}. 

A balanced distribution of authority supports human creativity but also preserves user agency and engagement \cite{muller_interactional_2023}. Authority allocation should be adaptive to user preferences and expertise level \cite{tsamados2024human}. For instance, experts often prefer to lead the creative process while relying on AI as an assistant \cite{biermann2022tool}. In contrast, novices benefit from AI taking a more directive role, helping structure their creative efforts \cite{dhillon2024shaping}. Given these differences, users value customizable authority settings that align with their AI literacy and creative workflow, ensuring optimal collaboration \cite{moruzzi_roles}. To maintain effective co-creativity, authority should shift dynamically based on task complexity, preventing AI from either becoming an overbearing decision-maker or an ineffective tool \cite{tsamados2024human}.

The distribution of authority significantly influences the creative process and outcomes. Research suggests that authority shifts dynamically across different phases of co-creation and contexts \cite{wan_it_2024}. 
In game design, increased AI authority can lead to user frustration when it overrides human intentions, demonstrating the importance of context-sensitive authority allocation \cite{larsson2022towards}. To prevent negative user experiences, co-creative AI should enable fluid authority transitions, allowing both human and AI inputs to meaningfully shape the creative output \cite{lobo_when_2024}.

The shift from supervisory control to human-machine teaming acknowledges the evolving role of AI in co-creativity, suggesting co-creative AI to be a co-equal partner \cite{rezwana2022designing}. 
Authority allocation determines whether AI functions as an assistant, co-creator, or independent agent, with increased AI reliability often leading to greater decision-making power for AI \cite{walsh2019effective}. Shared authority, where humans and AI collaboratively regulate the creative process, are emerging as a promising paradigm \cite{cimolino2022two}. 




\subsection{Strategies for Balancing Control}

\begin{figure}[h]
    \centering
    \includegraphics[width=0.55\columnwidth]{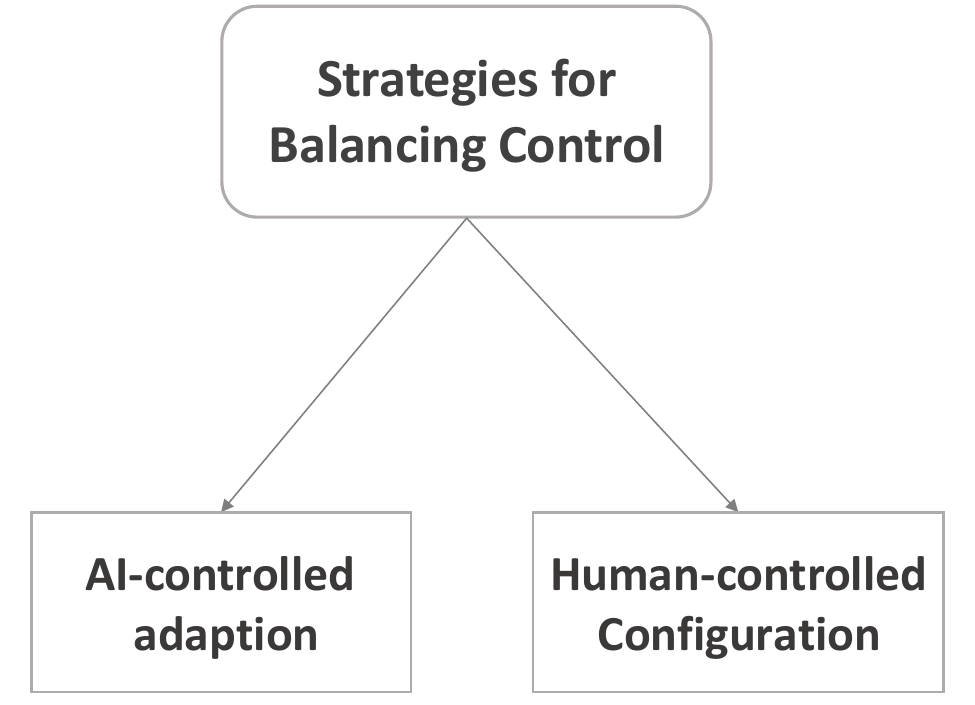}
    \caption{Strategies for balancing control in co-creativity}
    \label{strategy}
\end{figure}

As a part of MOSAAIC, in this section, we discuss the strategies to balance the control between humans and AI in co-creativity (Figure \ref{strategy}): \textit{AI-controlled adaptation} and \textit{human-controlled configuration}. The adoption of these strategies depends on the context, task type, user preference, and expertise~\cite{robot_control}. We discuss these strategies in detail below. 

\subsubsection{AI-Controlled Adaption}

AI-controlled adaptation to achieve dynamic control involves AI adjusting its autonomy, initiative, and authority based on the task and context. Based on the context, the system can adapt the dimensions of control, which has been explored in the domain of adaptive AI or variable autonomy. This is most notably exemplified by \textit{adaptive autonomous} systems that change their autonomy levels based on tasks executed by users~\cite{frasheri2017towards}. For instance, in team performance, the AI agent adjusting its autonomy based on the situation and delegating tasks and decision-making to the users improves human-AI team performance~\cite{salikutluk2024evaluation}. 


Another strategy of AI-controlled adaptation is AI systems monitoring the human's cognitive workload, emotional state, or expertise level to adjust their level of control~\cite{singh2022human}. The system might take on more control during routine tasks but cede control to the human during critical or complex situations~\cite{salikutluk2024evaluation}. For instance, if the human is experiencing high cognitive load, the AI might take on more responsibility to reduce the burden \cite{singh2022human}. Mixed-initiative systems, where both humans and AI can initiate changes, can also improve performance in navigation tasks and reduce operator workload \cite{lin2023beyond}. In this case, the AI is adapting its level of initiative based on the workload in other states of the co-creative process. This also shows the intertwined behavior of all three dimensions of control, where balancing control via AI adaptation involves AI systems adjusting their levels of autonomy, initiative, and authority according to the context so that they may engage with users in optimized control.

\subsubsection{Human-Controlled Configuration} 
Human-controlled configuration involves allowing users to customize and configure the control dynamics according to their preferences and goals. Preference varies significantly between expertise groups, suggesting the development of personalized co-creative systems \cite{lin_beyond_2023}. Customization, based on the user’s background, such as AI literacy, experience levels, and familiarity, is an important factor to consider in distributing control for co-creative systems~\cite{lin_beyond_2023}. 
Users' perceptions of an AI system are influenced by their creative goals and the system's ability to adapt to their expertise level \cite{lin_beyond_2023}. 

Humans can configure between different levels of AI involvement during the creative process to maximize system flexibility and enhance creative outcome generation \cite{dang2023choice}. This includes human configuration based on the creative process---the divergence stage of idea generation, and the convergence stage of evaluation and selection of ideas~\cite{creativephase}. For instance, in the divergence stage, experts often tend to prefer maintaining their autonomy when conceptualizing ideas~\cite{llm_video}, and in the convergence stage, novices prefer `in-action' feedback (feedback throughout a design task) to evaluate their creative session~\cite{feedback_2024}.

Another strategy of human configuration is utilizing steerable interfaces \cite{louie_expressive_2022} such as AI-steering tools (adjustable autonomy), example-based and semantic sliders (range of outcomes), and multiple alternatives of generated content to choose from (authority over outcomes)~\cite{novice_music_cocreation}. 
An important, yet overlooked, aspect of control is giving users multiple options to choose from and giving them a \textit{sense of control} \cite{leotti_born_2010}. Overall, balancing control via human-configuration involves allowing users to customize and configure their preferences and goals leading to a distribution of autonomy, initiative, and authority. 

\section{Case Studies}

We selected six co-creative systems representing different creative domains to analyze the distribution of control between humans and AI. Our goal was to validate MOSAAIC’s applicability across various co-creative domains while also examining and comparing trends in control distribution across domains. Convenience sampling was used to select systems from the literature seeded by systems the authors were familiar with.
Our sampling included the following domain-relevant co-creative systems: Cyborg \cite{branch_designing_2024}, LuminAI \cite{milka_luminai_2024}, Snake Story \cite{snakestory}, Reframer \cite{lawton_drawing_2023}, and Shimon \cite{gao2024music}. We also included ChatGPT \cite{openaiChatGPTOptimizing} to include one example of a general-purpose AI system that is used for co-creativity. 

Two researchers among the authors independently coded the six co-creative systems based on the system descriptions in their respective publications. They then met to identify and resolve discrepancies. Finally, disagreements were discussed and revised to reflect group consensus (see Table~\ref{tab:my_label}).

\begin{table}[h]
    \centering
    \small
    \begin{tabular}{|p{0.64in}|p{0.52in}|p{0.45in}|p{0.49in}|p{0.5in}|}
        \hline
        \textbf{System \& Domain} & \textbf{Autonomy} & \textbf{Initiative} & \textbf{Authority} & \textbf{Strategy} \\
        \hline
        \textbf{Cyborg} (Theatre) & Shared & Full-human & Full-human &  Human-controlled\\
        \hline
        \textbf{LuminAI} (Dance) & Shared & Shared & Shared & Both\\
        \hline
        \textbf{Snake Story} (Game-Design) & Shared & Shared & Full-human & Both\\
        \hline
        \textbf{ChatGPT} (Writing) & Shared & Full-human & Full-human & Both\\
        \hline
        \textbf{Reframer} (Drawing) & Full-human & Full-human & Full-human & Human-Controlled\\
        \hline
        \textbf{Shimon} (Music) & Shared & Shared & Shared & Both\\
        \hline
    \end{tabular}
    \caption{Analysis of the control distribution of six co-creative systems using MOSAAIC}
    \label{tab:my_label}
\end{table}

\subsection{Autonomy Distribution}
Systems were classified as having shared autonomy if both humans and AI had comparable abilities to make creative choices. If a system primarily followed user instructions with little creative autonomy, it was coded as full-human. Based on our analysis, five out of six systems exhibited shared autonomy, except for \textit{Reframer}, which generates images following the user prompts and instructions. While \textit{Reframer} allows users to control the level of variation in AI output, it does not have the ability to deviate from the prompt, making it more user-directed than co-creative. The remaining systems allow both humans and AI to make creative choices, leading to more emergent co-creation. For example, \textit{LuminAI}, an improvisational dance partner, generates dance movements independently but adapts to the human dancer’s motion, demonstrating shared autonomy. Similarly, \textit{ChatGPT} actively interprets user-provided prompts and contributes with its chosen creative strategies. 
\textit{Snake Story} is a co-creative game design system where the AI does not merely follow the user’s instructions but actively contributes by suggesting plot developments, steering the story in unexpected directions. \textit{Shimon} is an improvisational robotic marimba player that collaborates with human musicians by generating, adapting, and responding to musical phrases of the human partner in real time. \textit{Cyborg} is a system where AI generates text dialogues being enacted by a cyborg performer, but the generated dialogues are dependent on other actors' dialogues.

\subsection{Initiative Distribution}
Systems were classified as having shared initiative if both humans and AI had the ability to proactively contribute to the creative process. If a system primarily followed user input and instructions with a reactive response mode, it was coded as full-human. In our analysis of six systems, the dimension of initiative was equally split between shared initiative and full-human. We primarily identified this difference via proactivity and reactivity. We observed that co-creative systems such as \textit{Cyborg}, \textit{ChatGPT}, and \textit{Reframer} would only contribute to the creative process when the user instructs them and hence are `reactive' in the creative process. 
For instance, in \textit{Cyborg}, an AI-based improv theatre show, the LLM was prompted by speech recognition of the actors to generate the lines for the cyborg actor. This reactive approach distinguishes these systems as full-human initiative. On the other hand, systems with shared initiative involve both humans and AI proactively contributing to co-creation and do not wait for the human to begin the process. 
\textit{LuminAI} proactively starts to dance if the user does not start dancing within a threshold period during the co-creation; \textit{Shimon}, an improvisational robotic marimba player, takes initiative by autonomously generating musical phrases, setting the tempo, or leading a performance without requiring human input first; and \textit{Snake Story} collaboratively crafts a story between the user and the AI by interchanging the order contribution. 


\subsection{Authority Distribution}

Systems were classified as having shared authority if both humans and AI had similar abilities to make decisions and direct the co-creative process. This entails negotiation and mutual adaptation. Conversely, if a system primarily implemented the user's decision with little authority, it was coded as full-human. Based on our analysis, four out of six of the co-creative systems had full-human authority. In \textit{Cyborg}, a human curator reviews and selects AI-generated dialogues before they are performed by the cyborg. In \textit{Reframer}, users control the final output by accepting, rejecting, or modifying AI contributions using sliders and prompts. In \textit{Snake Story}, 
the AI lacks control over the narrative direction, as the human fully dictates the story’s progression and outcome. In \textit{ChatGPT}, the user determines the final goal, directs the flow of narration, and decides whether to include or exclude AI-generated content. What is particularly interesting about systems with shared authority, such as \textit{LuminAI} and \textit{Shimon}, is that both are improvisational AI without a fixed goal or tangible shared artifact. In these systems, the creative process itself is the product, meaning there is no final outcome over which authority must be exerted. \textit{LuminAI} exhibits shared authority as both the human and AI shape the dance interaction, with neither having full control over the creative direction. Similarly, \textit{Shimon} maintains shared authority by dynamically responding to human musicians.

\subsection{Strategies to Balance Control}


We identify strategies for modulating control between humans and AI in these six systems using MOSAAIC. Two out of the six systems have human-controlled configuration as the strategy to balance control. These are \textit{Cyborg} and \textit{Reframer}. Whereas four out of six, a majority of systems, have both AI-controlled adaptation and human-controlled configuration (classified as ``both'') as the strategies to balance control. These include \textit{LuminAI}, \textit{ChatGPT}, \textit{Shimon}, and \textit{Snake Story}. We will first describe the strategies for human-controlled configuration and then those for both. 

In \textit{Cyborg}, a human curator controls and refines AI-generated dialogues before they are enacted by the cyborg. The interface also allows the curator to guide the AI’s output by providing prompts such as ``be funnier," enabling them to adjust the AI's initiative and authority in the creative process. As an improv AI with the lack of a fixed goal, human-controlled configuration allows a well-balanced control dynamic that addresses the spontaneity that is embedded into the creative domain. On the other hand, in \textit{Reframer}, a drawing interface where there is a fixed goal, a human-controlled configuration would be implemented to provide variability on the control mechanisms for the creative autonomy of the AI generation. Likewise, initiative would also be dispersed by the human prompting the AI to initiative and authority to be distributed by negotiated decision-making. In this case, the user would be able to distribute autonomy, initiative, and authority in its drawing process upon directing its preference and goals. 

In shared strategies of control, the collective strategy of AI adaptation and human configuration would allow the system to adjust its autonomy, initiative, and authority based on context, tackling the user's nascent needs, while the human explicitly configures the system to their preferences and goals. In \textit{LuminAI} and \textit{Shimon}, this would allow all shared dimensions to evolve in unison, perhaps adding to different levels of shared control within the dimensions. \textit{LuminAI} adapts in real-time to a human dancer’s movements, while humans can configure movement parameters or styles to influence how the AI responds in the creative process. \textit{Shimon} modifies its musical phrases in response to human input, adjusting tempo and rhythm in real-time, while musicians retain control over aspects such as initiating musical themes or shaping compositional elements. 

Likewise, in \textit{ChatGPT} and \textit{Snake Story}, shared strategies would also balance control via the system's adaptation to the user's states (e.g., type of engagement), while the user equally adds their desired goals. \textit{ChatGPT} dynamically adjusts its responses based on the conversation history, while users fine-tune its output through iterative prompts and stylistic preferences. \textit{Snake Story} adapts autonomy by adjusting the level of AI-generated suggestions and initiative by varying its proactivity and reactivity based on the user's interaction frequency.

\section{Discussion and Conclusion}
 
The rapid rise and accessibility of GenAI technologies in creative domains are shifting agency from users to AI, transforming its role into agentic AI and reducing users' control over the creative process \cite{lin_beyond_2023}. Co-creativity literature emphasizes that while AI should have creative autonomy, users must also retain the ability to lead co-creation \cite{muller_interactional_2023}. Achieving this balance requires a control equilibrium between humans and AI, ensuring a harmonious co-creation.

In this paper, we introduce a framework for \textbf{managing optimization towards shared autonomy, authority, and initiative in co-creation (MOSAAIC)} that characterizes the key dimensions of control and strategies to balance control in human-AI co-creativity. Our framework emerges from a systematic literature review of 172 papers. The key dimensions of control identified in MOSAAIC are \textit{autonomy} (enabling the choice of creative action), \textit{initiative} (allowing the proactive contribution in co-creation), and \textit{authority} (granting the ability to decide and direct the creative process). While MOSAAIC's dimensions may be interconnected, each plays a distinct and vital role in defining control dynamics in human-AI co-creativity.
MOSAAIC identifies two strategies to balance control in co-creation, which are \textit{AI-controlled adaptation} and \textit{human-controlled configuration}. We also analyze six co-creative systems in different domains to show use cases of MOSAAIC and to provide how MOSAAIC attributes control distribution in co-creative AI systems of various domains.


\subsection{Use Cases of MOSAAIC}

By leveraging MOSAAIC’s strategies for optimizing control, researchers can identify ways to effectively balance control in human and AI co-creation. Below we discuss some use uses of MOSAAIC.

MOSAAIC can serve as a \textbf{tool for analyzing control distribution} in co-creative AI systems and can identify trends, strengths, and gaps in how control is allocated and optimized across co-creative AI systems. 
For instance, improvisational AI systems like LuminAI and Shimon in our case study, where the co-creative experience itself is the product rather than a tangible outcome, exhibit different control dynamics from more structured systems with varied combinations of the three dimensions. In these systems, all control dimensions were shared, unlike in other systems. 
Moreover, MOSAAIC also enables cross-domain comparisons beyond individual system analysis and serves as a benchmarking tool for evaluating co-creative AI across different creative fields. Evaluation tools and frameworks have been widely utilized in co-creativity domains \cite{kantosalo2020modalities,rezwana2022designing}. 

Second, MOSAAIC can be translated to a \textbf{configuration and customization tool}, enabling AI systems to adapt based on how users adjust control dimensions in co-creation. Research indicates that different user groups, such as experts and novices, have varying expectations from co-creative AI \cite{oh2018lead,moruzzi_roles}, and a configurable control system can accommodate these differences by providing greater adaptability. By allowing users to configure the dynamics of human-AI collaboration, MOSAAIC supports personalized experiences aligned with individual preferences or specific creative domain requirements, thus democratizing control in co-creative systems. This customization could be implemented in various ways, such as sliders for each control dimension, allowing users to shape the dynamics of human-AI control. By supporting such customization, MOSAAIC empowers users to shape their co-creative interactions, enhancing collaboration and creative synergy with the co-creative system. Since MOSAAIC is domain-agnostic, it can be applied across various creative domains. In our case study, we found that four out of the six systems incorporate both AI adaptation and human configuration for control optimization, highlighting a trend toward integrating AI adaptation alongside user customization.

Third, MOSAAIC can serve as a \textbf{tool for designing co-creative AI} with optimized control distribution by providing a structured characterization of control and strategies to balance control between humans and AI. Design frameworks are well-established in co-creativity research \cite{rezwana2022designing,kantosalo2020modalities}, aiding in the development of more effective systems. Our case study reveals opportunities to improve the design of co-creative AI utilizing balanced control. For example, three systems had shared initiative, while the rest were fully human-led, highlighting the need to explore initiative distribution—especially since most recent GenAI models remain reactive. In terms of authority, four out of six had full-human authority, indicating a scope to design systems with more dynamic authority.


\subsection{Limitations and Future Work}



The goal of MOSAAIC is to provide \textit{balance of control} and \textit{harmony} ~\cite{vinchon_artificial_2023} in the process of co-creation by addressing (1) the principle of maximal unobtrusiveness and (2) balance of interjection, as motivated in CC research. MOSAAIC currently identifies two high-level strategies for balancing control in co-creativity. However, how to decide exactly how to optimize a co-creative system in terms of control distribution is still an open question, except that we know that the adoption of strategies depends on the context, task type, user preference, and expertise~\cite{robot_control}. As such, in future work, we plan to identify context-appropriate strategies 
to optimize in different contexts and domains. Furthermore, although our literature review was comprehensive of ACC and ACM publications, we recognize that other relevant literature may have been omitted, and aim to expand our review to other libraries. We would expand our study towards identifying the factors that influence control, such as trust and ownership, to incorporate external factors that affect the balance of control in co-creation.

\bibliographystyle{iccc}
\bibliography{iccc}

\begin{thebibliography}{}

\bibitem[\protect\citeauthoryear{Bennett \bgroup et al.\egroup }{2023}]{bennett2023does}
Bennett, D.; Metatla, O.; Roudaut, A.; and Mekler, E.~D.
\newblock 2023.
\newblock How does hci understand human agency and autonomy?
\newblock In {\em Proceedings of the 2023 CHI},  1--18.

\bibitem[\protect\citeauthoryear{Biermann, Ma, and Yoon}{2022}]{biermann2022tool}
Biermann, O.~C.; Ma, N.~F.; and Yoon, D.
\newblock 2022.
\newblock From tool to companion: Storywriters want ai writers to respect their personal values and writing strategies.
\newblock In {\em Designing Interactive Systems Conference},  1209--1227.

\bibitem[\protect\citeauthoryear{Branch \bgroup et al.\egroup }{2024}]{branch_designing_2024}
Branch, B.; Mirowski, P.; Mathewson, K.; and Covaci, S. P.~A.
\newblock 2024.
\newblock Designing and {Evaluating} {Dialogue} {LLMs} for {Co}-{Creative} {Improvised} {Theatre}.

\bibitem[\protect\citeauthoryear{Chiou, Hawes, and Stolkin}{2021}]{mic}
Chiou, M.; Hawes, N.; and Stolkin, R.
\newblock 2021.
\newblock Mixed-initiative variable autonomy for remotely operated mobile robots.
\newblock {\em J. Hum.-Robot Interact.} 10(4).

\bibitem[\protect\citeauthoryear{Cimolino and Graham}{2022}]{cimolino2022two}
Cimolino, G., and Graham, T.~N.
\newblock 2022.
\newblock Two heads are better than one: A dimension space for unifying human and artificial intelligence in shared control.
\newblock In {\em Proceedings of the 2022 CHI},  1--21.

\bibitem[\protect\citeauthoryear{Dang \bgroup et al.\egroup }{2023}]{dang2023choice}
Dang, H.; Goller, S.; Lehmann, F.; and Buschek, D.
\newblock 2023.
\newblock Choice over control: How users write with large language models using diegetic and non-diegetic prompting.
\newblock In {\em Proceedings of CHI ’23},  1--17.

\bibitem[\protect\citeauthoryear{Davis \bgroup et al.\egroup }{2016}]{davis2016co}
Davis, N.; Hsiao, C.-P.; Singh, K.~Y.; and Magerko, B.
\newblock 2016.
\newblock Co-creative drawing agent with object recognition.
\newblock In {\em Proceedings of the AAAI Conference on Artificial Intelligence and Interactive Digital Entertainment}, volume~12,  9--15.

\bibitem[\protect\citeauthoryear{Davis \bgroup et al.\egroup }{2024}]{davis_five_2024}
Davis, N.; Deshpande, M.; Rezwana, J.; and Magerko, B.
\newblock 2024.
\newblock The {Five} {Pillars} of {Enaction} as a {Theoretical} {Framework} for {Co}-{Creative} {Artificial} {Intelligence}.

\bibitem[\protect\citeauthoryear{Dhillon \bgroup et al.\egroup }{2024}]{dhillon2024shaping}
Dhillon, P.~S.; Molaei, S.; Li, J.; Golub, M.; Zheng, S.; and Robert, L.~P.
\newblock 2024.
\newblock Shaping human-ai collaboration: varied scaffolding levels in co-writing with language models.
\newblock In {\em Proceedings of the 2024 CHI},  1--18.

\bibitem[\protect\citeauthoryear{E \bgroup et al.\egroup }{2024}]{feedback_2024}
E, J.~L.; Yen, Y.-C.~G.; Pan, I.~Y.; Lin, G.; Li, M.; Jin, H.; Chen, M.; Xia, H.; and Dow, S.~P.
\newblock 2024.
\newblock When to give feedback: Exploring tradeoffs in the timing of design feedback.
\newblock In {\em Proceedings of the 16th Conference on Creativity \& Cognition}, C\&C '24,  292–310.
\newblock New York, NY, USA: ACM.

\bibitem[\protect\citeauthoryear{Epstein \bgroup et al.\egroup }{2023}]{epstein2023art}
Epstein, Z.; Hertzmann, A.; of~Human~Creativity, I.; Akten, M.; Farid, H.; Fjeld, J.; Frank, M.~R.; Groh, M.; Herman, L.; Leach, N.; et~al.
\newblock 2023.
\newblock Art and the science of generative ai.
\newblock {\em Science} 380(6650):1110--1111.

\bibitem[\protect\citeauthoryear{Feldman}{2017}]{feldman_co-creation_2017}
Feldman, S.~S.
\newblock 2017.
\newblock Co-{Creation}: {Human} and {AI} {Collaboration} in {Creative} {Expression}.
\newblock In {\em Electronic Visualisation and the Arts (EVA 2017)}.
\newblock BCS Learning \& Development.

\bibitem[\protect\citeauthoryear{Ferrario, Facchini, and Termine}{2024}]{ferrario2024experts}
Ferrario, A.; Facchini, A.; and Termine, A.
\newblock 2024.
\newblock Experts or authorities? the strange case of the presumed epistemic superiority of artificial intelligence systems.
\newblock {\em Minds and Machines} 34(3):30.

\bibitem[\protect\citeauthoryear{Frasheri, {\c{C}}{\"u}r{\"u}kl{\"u}, and Ekstr{\"o}m}{2017}]{frasheri2017towards}
Frasheri, M.; {\c{C}}{\"u}r{\"u}kl{\"u}, B.; and Ekstr{\"o}m, M.
\newblock 2017.
\newblock Towards collaborative adaptive autonomous agents.
\newblock In {\em ICAART (1)}.

\bibitem[\protect\citeauthoryear{Gao \bgroup et al.\egroup }{2024}]{gao2024music}
Gao, X.; Rogel, A.; Sankaranarayanan, R.; Dowling, B.; and Weinberg, G.
\newblock 2024.
\newblock Music, body, and machine: gesture-based synchronization in human-robot musical interaction.
\newblock {\em Frontiers in Robotics and AI} 11:1461615.

\bibitem[\protect\citeauthoryear{Guzman and Lewis}{2020}]{guzman2020artificial}
Guzman, A.~L., and Lewis, S.~C.
\newblock 2020.
\newblock Artificial intelligence and communication: A human--machine communication research agenda.
\newblock {\em New media \& society} 22(1):70--86.

\bibitem[\protect\citeauthoryear{Hardin and Goodrich}{2009}]{robot_control}
Hardin, B., and Goodrich, M.~A.
\newblock 2009.
\newblock On using mixed-initiative control: a perspective for managing large-scale robotic teams.
\newblock In {\em Proceedings of the 4th ACM/IEEE International Conference on Human Robot Interaction}, HRI '09,  165–172.
\newblock ACM.

\bibitem[\protect\citeauthoryear{Heer}{2019}]{heer2019agency}
Heer, J.
\newblock 2019.
\newblock Agency plus automation: Designing artificial intelligence into interactive systems.
\newblock {\em Proceedings of the National Academy of Sciences} 116(6):1844--1850.

\bibitem[\protect\citeauthoryear{Kabir}{2024}]{kabir_unleash_2024}
Kabir, M.~N.
\newblock 2024.
\newblock View of {Unleashing} {Human} {Potential}: {A} {Framework} for {Augmenting} {Co}-{Creation} with {Generative} {AI}.

\bibitem[\protect\citeauthoryear{Kantosalo \bgroup et al.\egroup }{2014}]{kantosalo_isolation_2014}
Kantosalo, A.; Toivanen, J.~M.; Xiao, P.; and Toivonen, H.
\newblock 2014.
\newblock From {Isolation} to {Involvement}: {Adapting} {Machine} {Creativity} {Software} to {Support} {Human}-{Computer} {Co}-{Creation}.

\bibitem[\protect\citeauthoryear{Kantosalo \bgroup et al.\egroup }{2020}]{kantosalo2020modalities}
Kantosalo, A.; Ravikumar, P.~T.; Grace, K.; and Takala, T.
\newblock 2020.
\newblock Modalities, styles and strategies: An interaction framework for human-computer co-creativity.
\newblock In {\em International Conference on Computational Creativity},  57--64.

\bibitem[\protect\citeauthoryear{Kim \bgroup et al.\egroup }{2023}]{kim2023one}
Kim, T.; Molina, M.~D.; Rheu, M.; Zhan, E.~S.; and Peng, W.
\newblock 2023.
\newblock One ai does not fit all: A cluster analysis of the laypeople’s perception of ai roles.
\newblock In {\em Proceedings of the 2023 CHI},  1--20.

\bibitem[\protect\citeauthoryear{Koch, Ravikumar, and Calegario}{2021}]{koch_agency_2021}
Koch, J.; Ravikumar, P.~T.; and Calegario, F.
\newblock 2021.
\newblock Agency in {Co}-{Creativity}: {Towards} a {Structured} {Analysis} of a {Concept}.
\newblock In Garza, A. G. d.~S.; Veale, T.; Aguilar, W.; and Pérez, R. P.~y., eds., {\em Proceedings of the {Twelfth} {International} {Conference} on {Computational} {Creativity}, {September} 14-18, 2021},  449--452.
\newblock Association for Computational Creativity.

\bibitem[\protect\citeauthoryear{Larsson, Font, and Alvarez}{2022}]{larsson2022towards}
Larsson, T.; Font, J.; and Alvarez, A.
\newblock 2022.
\newblock Towards ai as a creative colleague in game level design.
\newblock In {\em Proceedings of the AAAI Conference on Artificial Intelligence and Interactive Digital Entertainment}, volume~18,  137--145.

\bibitem[\protect\citeauthoryear{Lawton \bgroup et al.\egroup }{2023}]{lawton_drawing_2023}
Lawton, T.; Ibarrola, F.~J.; Ventura, D.; and Grace, K.
\newblock 2023.
\newblock Drawing with {Reframer}: {Emergence} and {Control} in {Co}-{Creative} {AI}.
\newblock In {\em Proceedings of the 28th IUI},  264--277.
\newblock Sydney NSW Australia: ACM.

\bibitem[\protect\citeauthoryear{Leotti, Iyengar, and Ochsner}{2010}]{leotti_born_2010}
Leotti, L.~A.; Iyengar, S.~S.; and Ochsner, K.~N.
\newblock 2010.
\newblock Born to {Choose}: {The} {Origins} and {Value} of the {Need} for {Control}.
\newblock {\em Trends in cognitive sciences} 14(10):457--463.

\bibitem[\protect\citeauthoryear{Liapis, Yannakakis, and Togelius}{2014}]{liapis2014computational}
Liapis, A.; Yannakakis, G.~N.; and Togelius, J.
\newblock 2014.
\newblock Computational game creativity.
\newblock ICCC.

\bibitem[\protect\citeauthoryear{Lin \bgroup et al.\egroup }{2023a}]{lin_beyond_2023}
Lin, Z.; Ehsan, U.; Agarwal, R.; Dani, S.; Vashishth, V.; and Riedl, M.
\newblock 2023a.
\newblock Beyond {Prompts}: {Exploring} the {Design} {Space} of {Mixed}-{Initiative} {Co}-{Creativity} {Systems}.

\bibitem[\protect\citeauthoryear{Lin \bgroup et al.\egroup }{2023b}]{lin2023beyond}
Lin, Z.; Ehsan, U.; Agarwal, R.; Dani, S.; Vashishth, V.; and Riedl, M.
\newblock 2023b.
\newblock Beyond prompts: Exploring the design space of mixed-initiative co-creativity systems.
\newblock {\em arXiv preprint arXiv:2305.07465}.

\bibitem[\protect\citeauthoryear{Liu, Huang, and Holopainen}{2023}]{liu2023use}
Liu, H.~X.; Huang, Y.; and Holopainen, J.
\newblock 2023.
\newblock How to use generative ai as a design material for future human-computer (co-) creation?

\bibitem[\protect\citeauthoryear{Lobo \bgroup et al.\egroup }{2024}]{lobo_when_2024}
Lobo, I.; Koch, J.; Renoux, J.; Batina, I.; and Prada, R.
\newblock 2024.
\newblock When {Should} {I} {Lead} or {Follow}: {Understanding} {Initiative} {Levels} in {Human}-{AI} {Collaborative} {Gameplay}.
\newblock In {\em Proceedings of the 2024 {ACM} {Designing} {Interactive} {Systems} {Conference}}, {DIS} '24,  2037--2056.
\newblock New York, NY, USA: ACM.

\bibitem[\protect\citeauthoryear{Louie \bgroup et al.\egroup }{2020}]{novice_music_cocreation}
Louie, R.; Coenen, A.; Huang, C.~Z.; Terry, M.; and Cai, C.~J.
\newblock 2020.
\newblock Novice-ai music co-creation via ai-steering tools for deep generative models.
\newblock In {\em Proceedings of CHI ’20}, CHI '20,  1–13.
\newblock New York, NY, USA: ACM.

\bibitem[\protect\citeauthoryear{Louie, Engel, and Huang}{2022}]{louie_expressive_2022}
Louie, R.; Engel, J.; and Huang, C.-Z.~A.
\newblock 2022.
\newblock Expressive communication: Evaluating developments in generative models and steering interfaces for music creation.
\newblock In {\em Proceedings of the 27th International Conference on Intelligent User Interfaces}, IUI '22.
\newblock New York, NY, USA: ACM.

\bibitem[\protect\citeauthoryear{Lubart}{2005}]{lubart_how_2005}
Lubart, T.
\newblock 2005.
\newblock How can computers be partners in the creative process: {Classification} and commentary on the {Special} {Issue}.
\newblock {\em International Journal of Human-Computer Studies} 63(4):365--369.

\bibitem[\protect\citeauthoryear{Maher}{2012}]{maher_who_2012}
Maher, M.
\newblock 2012.
\newblock {Computational} and {Collective} {Creativity}: {Who} is {Being} {Creative?}
\newblock In {\em In Proceedings of The Third International Conference on Computational Creativity, University College Dublin},  67--71.
\newblock Association for Computational Creativity.

\bibitem[\protect\citeauthoryear{Manovich and Arielli}{2024}]{manovich_2024}
Manovich, L., and Arielli, E.
\newblock 2024.
\newblock {\em Artificial Aesthetics: Generative AI, Art and Visual Media}.

\bibitem[\protect\citeauthoryear{Margarido \bgroup et al.\egroup }{2024}]{margarido_boosting_2024}
Margarido, S.; Roque, L.; Machado, P.; and Martins, P.
\newblock 2024.
\newblock Boosting {Mixed}-{Initiative} {Co}-{Creativity} in {Game} {Design}: {A} {Tutorial}.
\newblock arXiv:2401.05999 [cs].

\bibitem[\protect\citeauthoryear{Margarido \bgroup et al.\egroup }{2025}]{margarido_mi-ccy_2025}
Margarido, S.; Roque, L.; Machado, P.; and Martins, P.
\newblock 2025.
\newblock {MI}-{CCy} {Quantifier}: {A} {Framework} for {Quantifying} {Mixed}-{Initiative} {Co}-creativity in {Human}-{AI} {Collaborations}.
\newblock In {\em Progress in {Artificial} {Intelligence}},  3--15.
\newblock Springer, Cham.
\newblock ISSN: 1611-3349.

\bibitem[\protect\citeauthoryear{Meta}{2024}]{MovieGen}
Meta.
\newblock 2024.
\newblock {Movie} {Gen}: {A} {Cast} of {Media} {Foundation} {Models}.
\newblock AI at Meta.
\newblock https://ai.meta.com/static-resource/movie-gen-research-paper.

\bibitem[\protect\citeauthoryear{Morris \bgroup et al.\egroup }{2024}]{morris_levels_2024}
Morris, M.~R.; Sohl-dickstein, J.; Fiedel, N.; Warkentin, T.; Dafoe, A.; Faust, A.; Farabet, C.; and Legg, S.
\newblock 2024.
\newblock Levels of {AGI} for {Operationalizing} {Progress} on the {Path} to {AGI}.
\newblock arXiv:2311.02462 [cs].

\bibitem[\protect\citeauthoryear{Moruzzi and Margarido}{2024a}]{moruzzi_customizing_2024}
Moruzzi, C., and Margarido, S.
\newblock 2024a.
\newblock Customizing the {Balance} between {User} and {System} {Agency} in {Human}-{AI} {Co}-{Creative} {Processes}.

\bibitem[\protect\citeauthoryear{Moruzzi and Margarido}{2024b}]{moruzzi_roles}
Moruzzi, C., and Margarido, S.
\newblock 2024b.
\newblock A user-centered framework for human-ai co-creativity.
\newblock In {\em Proceedings of CHI ’24 EA}, CHI EA '24.
\newblock New York, NY, USA: ACM.

\bibitem[\protect\citeauthoryear{Moruzzi}{2022}]{moruzzi2022creative}
Moruzzi, C.
\newblock 2022.
\newblock Creative agents: rethinking agency and creativity in human and artificial systems.
\newblock {\em Journal of Aesthetics and Phenomenology} 9(2):245--268.

\bibitem[\protect\citeauthoryear{Muller, Candello, and Weisz}{2023}]{muller_interactional_2023}
Muller, M.; Candello, H.; and Weisz, J.
\newblock 2023.
\newblock Interactional {Co}-{Creativity} of {Human} and {AI} in {Analogy}-{Based} {Design}.

\bibitem[\protect\citeauthoryear{Muyskens \bgroup et al.\egroup }{2024}]{muyskens2024can}
Muyskens, K.; Ma, Y.; Menikoff, J.; Hallinan, J.; and Savulescu, J.
\newblock 2024.
\newblock When can we kick (some) humans “out of the loop”? an examination of the use of ai in medical imaging for lumbar spinal stenosis.
\newblock {\em Asian Bioethics Review}.

\bibitem[\protect\citeauthoryear{Oh \bgroup et al.\egroup }{2018}]{oh2018lead}
Oh, C.; Song, J.; Choi, J.; Kim, S.; Lee, S.; and Suh, B.
\newblock 2018.
\newblock I lead, you help but only with enough details: Understanding user experience of co-creation with artificial intelligence.
\newblock In {\em Proceedings of the 2018 CHI conference on human factors in computing systems},  1--13.

\bibitem[\protect\citeauthoryear{OpenAI}{2022}]{openaiChatGPTOptimizing}
OpenAI.
\newblock 2022.
\newblock {C}hat{G}{P}{T}: {O}ptimizing {L}anguage {M}odels for {D}ialogue --- openai.com.

\bibitem[\protect\citeauthoryear{Pease, Colton, and Banar}{2023}]{pease2023notion}
Pease, A.; Colton, S.; and Banar, B.
\newblock 2023.
\newblock On the notion of creative personhood.
\newblock In {\em 14th International Conference on Computational Creativity},  117--121.
\newblock Association for Computational Creativity.

\bibitem[\protect\citeauthoryear{Rezwana and Maher}{2022}]{rezwana2022designing}
Rezwana, J., and Maher, M.~L.
\newblock 2022.
\newblock Designing creative ai partners with cofi: A framework for modeling interaction in human-ai co-creative systems.
\newblock {\em ACM Transactions on Computer-Human Interaction}.

\bibitem[\protect\citeauthoryear{Salikutluk \bgroup et al.\egroup }{2024}]{salikutluk2024evaluation}
Salikutluk, V.; Sch{\"o}pper, J.; Herbert, F.; Scheuermann, K.; Frodl, E.; Balfanz, D.; J{\"a}kel, F.; and Koert, D.
\newblock 2024.
\newblock An evaluation of situational autonomy for human-ai collaboration in a shared workspace setting.
\newblock In {\em Proceedings of CHI ’24},  1--17.

\bibitem[\protect\citeauthoryear{Shaer \bgroup et al.\egroup }{2024}]{creativephase}
Shaer, O.; Cooper, A.; Mokryn, O.; Kun, A.~L.; and Ben~Shoshan, H.
\newblock 2024.
\newblock Ai-augmented brainwriting: Investigating the use of llms in group ideation.
\newblock In {\em Proceedings of CHI ’24}, CHI '24.
\newblock New York, NY, USA: ACM.

\bibitem[\protect\citeauthoryear{Singh and Heard}{2022}]{singh2022human}
Singh, S., and Heard, J.
\newblock 2022.
\newblock Human-aware reinforcement learning for adaptive human robot teaming.
\newblock In {\em 2022 17th ACM/IEEE International Conference on Human-Robot Interaction (HRI)},  1049--1052.
\newblock IEEE.

\bibitem[\protect\citeauthoryear{Stuart}{2003}]{stuart2003russell}
Stuart, J.
\newblock 2003.
\newblock Russell and peter norvig.
\newblock {\em Artificial Intelligence: A Modern Approach, Prentice Hall}.

\bibitem[\protect\citeauthoryear{Trajkova \bgroup et al.\egroup }{2024}]{milka_luminai_2024}
Trajkova, M.; Long, D.; Deshpande, M.; Knowlton, A.; and Magerko, B.
\newblock 2024.
\newblock Exploring collaborative movement improvisation towards the design of luminai—a co-creative ai dance partner.
\newblock In {\em Proceedings of the 2024 CHI}.
\newblock New York, NY, USA: Association for Computing Machinery.

\bibitem[\protect\citeauthoryear{Tsamados, Floridi, and Taddeo}{2024}]{tsamados2024human}
Tsamados, A.; Floridi, L.; and Taddeo, M.
\newblock 2024.
\newblock Human control of ai systems: from supervision to teaming.
\newblock {\em AI and Ethics}  1--14.

\bibitem[\protect\citeauthoryear{Vinchon \bgroup et al.\egroup }{2023}]{vinchon_artificial_2023}
Vinchon, F.; Lubart, T.; Bartolotta, S.; Gironnay, V.; Botella, M.; Bourgeois-Bougrine, S.; Burkhardt, J.; Bonnardel, N.; Corazza, G.~E.; Glăveanu, V.; Hanchett~Hanson, M.; Ivcevic, Z.; Karwowski, M.; Kaufman, J.~C.; Okada, T.; Reiter‐Palmon, R.; and Gaggioli, A.
\newblock 2023.
\newblock Artificial {Intelligence} \& {Creativity}: {A} {Manifesto} for {Collaboration}.
\newblock {\em The Journal of Creative Behavior} 57(4):472--484.

\bibitem[\protect\citeauthoryear{Walsh \bgroup et al.\egroup }{2019}]{walsh2019effective}
Walsh, T.; Levy, N.; Bell, G.; Elliott, A.; Maclaurin, J.; Mareels, I.; and Wood, F.
\newblock 2019.
\newblock {\em The effective and ethical development of artificial intelligence: an opportunity to improve our wellbeing}.
\newblock Australian Council of Learned Academia.

\bibitem[\protect\citeauthoryear{Wan \bgroup et al.\egroup }{2024}]{wan_it_2024}
Wan, Q.; Hu, S.; Zhang, Y.; Wang, P.; Wen, B.; and Lu, Z.
\newblock 2024.
\newblock "{It} {Felt} {Like} {Having} a {Second} {Mind}": {Investigating} {Human}-{AI} {Co}-creativity in {Prewriting} with {Large} {Language} {Models}.
\newblock {\em Proc. ACM Hum.-Comput. Interact.} 8(CSCW1):84:1--84:26.

\bibitem[\protect\citeauthoryear{Wang \bgroup et al.\egroup }{2024a}]{llm_video}
Wang, B.; Li, Y.; Lv, Z.; Xia, H.; Xu, Y.; and Sodhi, R.
\newblock 2024a.
\newblock Lave: Llm-powered agent assistance and language augmentation for video editing.
\newblock In {\em Proceedings of the 29th International Conference on Intelligent User Interfaces}, IUI '24,  699–714.
\newblock New York, NY, USA: ACM.

\bibitem[\protect\citeauthoryear{Wang \bgroup et al.\egroup }{2024b}]{promptcharm}
Wang, Z.; Huang, Y.; Song, D.; Ma, L.; and Zhang, T.
\newblock 2024b.
\newblock Promptcharm: Text-to-image generation through multi-modal prompting and refinement.
\newblock In {\em Proceedings of CHI ’24}, CHI '24.
\newblock New York, NY, USA: ACM.

\bibitem[\protect\citeauthoryear{Wu \bgroup et al.\egroup }{2021}]{AI_creativity}
Wu, Z.; Ji, D.; Yu, K.; Zeng, X.; Wu, D.; and Shidujaman, M.
\newblock 2021.
\newblock {\em AI Creativity and the Human-AI Co-creation Model}.
\newblock IEEE.
\newblock  171--190.

\bibitem[\protect\citeauthoryear{Yang \bgroup et al.\egroup }{2024}]{snakestory}
Yang, D.; Kleinman, E.; Troiano, G.~M.; Tochilnikova, E.; and Harteveld, C.
\newblock 2024.
\newblock Snake story: Exploring game mechanics for mixed-initiative co-creative storytelling games.
\newblock In {\em Proceedings of the 19th International Conference on the Foundations of Digital Games}, FDG '24.
\newblock ACM.

\bibitem[\protect\citeauthoryear{Yannakakis, Liapis, and Alexopoulos}{2014}]{yannakakis_mixed-initiative_2014}
Yannakakis, G.~N.; Liapis, A.; and Alexopoulos, C.
\newblock 2014.
\newblock Mixed-{Initiative} {Co}-{Creativity}.

\bibitem[\protect\citeauthoryear{Yao \bgroup et al.\egroup }{2024}]{yao_human_2024}
Yao, F.; Li, C.; Nekipelov, D.; Wang, H.; and Xu, H.
\newblock 2024.
\newblock Human vs. {Generative} {AI} in {Content} {Creation} {Competition}: {Symbiosis} or {Conflict}?
\newblock arXiv:2402.15467 [cs].

\end{thebibliography}

\end{document}